\RequirePackage{silence}
\documentclass[runningheads]{llncs}

\usepackage{amsmath,amssymb,amsfonts}
\usepackage{graphicx}
\usepackage{pifont}

\usepackage{xcolor}
\usepackage[figuresright]{rotating}

\usepackage{graphicx}
\graphicspath{{/}{fig/}}

\usepackage{array}
\usepackage{textcomp}
\usepackage{xcolor}
\usepackage{multirow}
\usepackage{booktabs}

\usepackage{mathtools}
\usepackage{breqn}
\usepackage{float}
\usepackage{enumerate}

\usepackage{pgfplots}
\pgfplotsset{compat=1.7}
\usepgfplotslibrary{groupplots}

\usepackage{caption}
\usepackage{subcaption}

\usepackage{multirow,tabularx}
\usepackage{hyperref}
\usepackage{flushend}
\usepackage{algorithmic}
\usepackage{listings}
\usepackage[vlined, ruled, shortend]{algorithm2e}
\newlength\figureheight
\newlength\figurewidth
\SetAlCapNameFnt{\footnotesize}
\SetAlCapFnt{\footnotesize}

\begin{document}
\title{Decentralized Vision-Based Byzantine Agent Detection in Multi-Robot Systems with IOTA Smart Contracts}
\author{
    Sahar Salimpour \and
    Farhad Keramat \and \\
    Jorge Peña Queralta \and 
    Tomi Westerlund
}%
\authorrunning{Sahar Salimpour et al.}
\titlerunning{Decentralized Vision-Based Byzantine Agent Detection with IOTA}
\institute{
        Turku Intelligent Embedded and Robotic Systems \\
        University of Turku, Finland\\ 
        \email{\{sahars, fakera, jopequ, tovewe\}@utu.fi} \\
        \url{https://tiers.utu.fi}
    }
\maketitle
%
%
%
%%%%%%%%%%%%%%%%%%%%%%%%%%%%%%%%%%%%%%%%%%%%%%
%%                                          %%
%%                SECTIONS                  %%
%%                                          %%
%%%%%%%%%%%%%%%%%%%%%%%%%%%%%%%%%%%%%%%%%%%%%%
\begin{abstract}

    Multiple opportunities lie at the intersection of multi-robot systems and distributed ledger technologies (DLTs). In this work, we investigate the potential of new DLT solutions such as IOTA, for detecting anomalies and byzantine agents in multi-robot systems in a decentralized manner. Traditional blockchain approaches are not applicable to real-world networked and decentralized robotic systems where connectivity conditions are not ideal. To address this, we leverage recent advances in partition-tolerant and byzantine-tolerant collaborative decision-making processes with IOTA smart contracts. We show how our work in vision-based anomaly and change detection can be applied to detecting byzantine agents within multiple robots operating in the same environment. We show that IOTA smart contracts add a low computational overhead while allowing to build trust within the multi-robot system. The proposed approach effectively enables byzantine robot detection based on the comparison of images submitted by the different robots and detection of anomalies and changes between them.
    

\end{abstract}
%
%
% \author{Sahar Salimpour\inst{1} \and
% Farhad Keramat\inst{1} \and
% Jorge Pe\~na Queralta\inst{1} \and
% Tomi Westerlund\inst{1}}
% %                                                                         %
% \authorrunning{S. Salimpour et al.}
% \titlerunning{Distributed Vision-Based Byzantine Agent Detection}
% %                                                                               
% \institute{Turku Intelligent Embedded and Robotic Systems (TIERS) Lab, University of Turku, Finland\\
% \email{\{sahars, fakera, jopequ, tovewe\}@utu.fi}}
%                                                                               
% \maketitle       

%%%%%%%%%%%%%%%%%%%%%%%%%%%%%%%%%%%%%%%%%%%%%%
%%                                          %%
%%              INTRODUCTION                %%
%%                                          %%
%%%%%%%%%%%%%%%%%%%%%%%%%%%%%%%%%%%%%%%%%%%%%%

\section{Introduction}\label{sec:introduction}

In recent years, byzantine agent detection has become an important aspect of distributed autonomous systems~\cite{ferrer2021following,keramat2022partition,strobel2020blockchain}. Indeed, with the growth and increasing ubiquity of autonomous robots, security and safety issues for systems deployed in the real world have attracted an ever-growing attention in both industrial and academic areas~\cite{queralta2022secure,queralta2020enhancing}. As robotic systems are deployed in larger numbers, single autonomous robots have been replaced by fleets of multi-robot systems that need to coordinate and collaborate. Many multi-robot applications, including security monitoring, public safety~\cite{xiao2021blockchain}, industrial applications~\cite{salimi2022secure,salimi2022towards}, and Internet of Things (IoT) systems~\cite{mohanta2020survey}, are at risk of being manipulated through the injection of fabricated or noisy data, or the performance of a large system might significantly decrease because of a single malicious or byzantine actor. Consequently, byzantine robots could potentially lead to a failure of the entire multi-robot operation. Therefore, it is important to be able to detect and neutralize the actions of byzantine agents, particularly if operating in environments together with humans.

\begin{figure}[t]
    \centering
    \includegraphics[width=\textwidth]{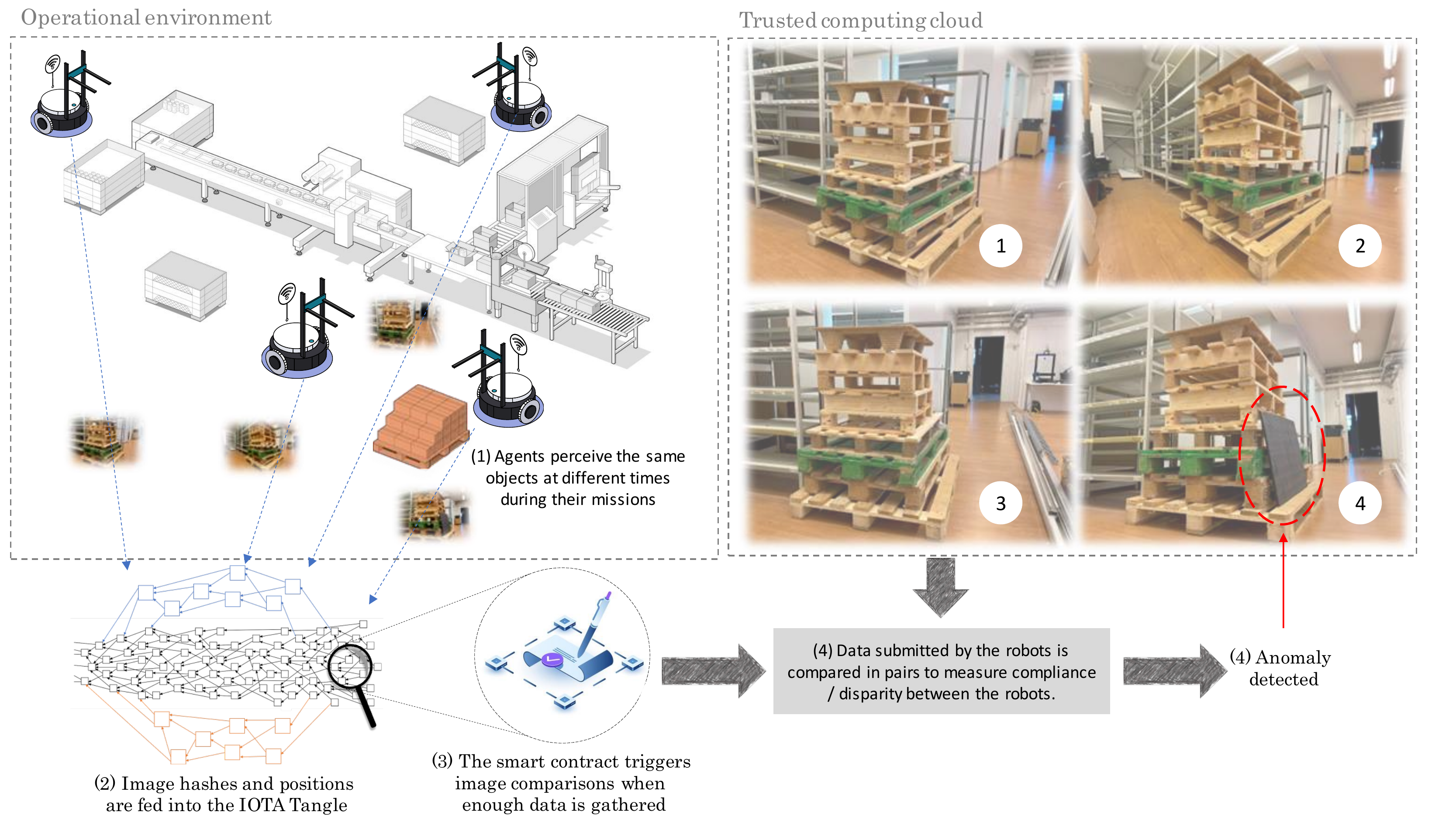}
    \caption{Conceptual illustration of the proposed vision-based byzantine agent detection approach with IOTA smart contracts.}
    \label{fig:concep}
\end{figure}

In multi-robot systems, vision-based perception often plays a major role in use cases involving safety, surveillance, and environment monitoring. Vision-based approaches to detect changes or anomalies in the environment can potentially be used to also detect differences between sensing data gathered by different robots operating in a common environment. A majority of visual anomaly detection problems are focused on a specific class of images and attempt to identify pixel-level anomalies in them. They mostly require training their deep learning (DL)-based models using large amount of \textit{normal} data~\cite{di2021pixel,wang2021cognitive,wang2022deep}. However, the detection of anomalies based on visual data in more general and potentially unknown environments easily becomes a challenging task, especially in the context of mobile robotic applications.

Novel approaches in the literature with potential to address the identification of byzantine robots in multi-robot systems are blockchain-based solutions through smart contracts. Blockchain technology was originally developed for the purposes of financial transactions~\cite{nakamoto2008bitcoin}, but it has also been utilized as a distributed computing framework for applications in general, e.g., within the Internet of Things (IoT) domain, as well as in multi-robot systems. A distributed system integrating blockchain technology is a priori capable of delivering a trusted and decentralized system between independent and untrusted agents. In the case of autonomous robots, this allows for decentralized collaborative decision making without the need for a third-party central organization. By doing so, a consistent global state makes the whole system resilient and fault-tolerant against byzantine robots.

IOTA smart contracts, designed for IoT devices, are one of the promising distributed ledger technology (DLT) solutions that can be used in multi-robot systems. In our previous work~\cite{keramat2022partition}, we have presented a general partition-tolerant and byzantine-tolerant framework built on top of IOTA smart contracts and integrated to ROS\,2. By leveraging this framework, all non-byzantine robots could reach a consensus about which robot is byzantine in a decentralized manner.

Blockchains or other distributed ledger technologies (DLTs) have potential to be an innovative solution to vision applications. However, to the best of our knowledge, no studies have been conducted on this topic within the context of multi-robot systems. In this study, we present a framework to detect byzantine robot(s) in a secure network and operating in a common environment by analyzing the RGB images which are captured by each robot. In the proposed method we use our previous study~\cite{salimpour2022self}, presenting a general framework to detect anomalies and changes between images, to compare in pairs images gathered by different robots. In this case, an anomaly could be something that has been moved or removed from the environment or something that does not belong there, as well as potentially altered or fabricated data.

This paper therefore integrates a vision-based approach for anomaly and change detection in autonomous inspection robots together with IOTA smart contracts, as illustrated by Fig.~\ref{fig:concep}. The result is a decentralized solution to anomaly detection that can be applied to byzantine agent detection within multi-robot systems. The blockchain serves as a tool for storing agent locations and image hashes, while smart contracts calculate where and when to perform the anomaly and change detection once enough data has been acquired. The DL model itself runs on a trusted server, owing to the impossibility of integrating such complex computation (deep neural networks) within a smart contract, and therefore limiting the decentralization of the solution. However, this is a first step towards a fully distributed implementation where multiple nodes will be able to validate the output of the DL models. In summary, the main contributions of this work are the following:%
\begin{enumerate}[i)]
    \item The design and implementation of a blockchain-based approach to byzantine agent detection using IOTA smart contract and vision sensors;
    \item the extension of our previous work in anomaly and change detection for autonomous inspection robots to comparing data from multiple robots operating in the same environment towards byzantine agent detection; and
    \item the integration of the DL models with IOTA smart contracts that trigger data comparisons after tracking the position of robots and the location of gathered data.
\end{enumerate}

The rest of the manuscript is structured as follows. Section II discusses related research on blockchain technology in robotic systems, and the problem of anomaly detection in multi-robot systems. A general introduction on the background is given in section III, and a description of our methodological approach is provided in section IV. Section V presents the results, and Section VI summarizes the work and points to future directions.

%%%%%%%%%%%%%%%%%%%%%%%%%%%%%%%%%%%%%%%%%%%%%%
%%                                          %%
%%              RELATED WORKS               %%
%%                                          %%
%%%%%%%%%%%%%%%%%%%%%%%%%%%%%%%%%%%%%%%%%%%%%%

% \newpage
\section{Related Work} \label{sec:related_work}

Generally, byzantine and fault detection in robotics can be divided into two major groups: self-monitoring and group-monitoring anomaly detection. Several studies proposed the self-monitoring approach, in which each robot detects anomalies independently. A framework for detecting mechanical faults and sensor faults in wheeled robots was presented in~\cite{yuan2015novel}. Tingting et al~\cite{chen2020unsupervised}. proposed an unsupervised anomaly detection model using a sliding-window convolutional variational autoencoder in terms of time series effect. However, swarm-level anomaly detection methods analyze the collaboration between robots and the data collected from the entire swarm~\cite{miller2021survey}.

Many studies have addressed byzantine robot detection in multi-robot missions using blockchain technology. In~\cite{strobel2018managing} a blockchain-based approach was explored for swarm robotics systems with byzantine robots. The authors utilized Ethereum-based decentralized smart contracts to detect and remove the byzentine swarm members. Their approach was evaluated using a collective decision-making scenario in which robots must agree on the most frequent tile color in an environment. 
In another work~\cite{ferrer2021following}, a blockchain was used as a secure communication tool in Byzantine Follow The Leader (BFTL) missions. Through their approach, leader robots guide follower robots to specific destinations under the threat of Byzantine robots misdirecting them. In addition, some research conducted to implement blockchain protocols into secure communication multi-agent systems with unmanned aerial vehicles~\cite{santos2021towards,kapitonov2017blockchain}.
 
The immutability of blockchain makes it a secure solution for detecting anomalous behaviors and attacks in various systems with chains of information blocks, such as industrial control systems~\cite{jadidi2020securing}, electricity consumption~\cite{li2020blockchain}, and health systems~\cite{castaldo2018blockchain}. The authors in~\cite{lopes2019detecting} implemented smart contracts to store robot information and compute them to detect anomalies, which were simulated internal failure, in machinery and register them in the blockchain. Golomb et al.~\cite{golomb2018ciota} introduced a collaborative anomaly detection model for a large network of IoT devices by leveraging blockchain technology in conjunction with extensible Markov model.

A number of recent studies have focused on detecting visual anomalies within specific image classes, such as railway images~\cite{wang2022deep}, road datasets~\cite{santhosh2020anomaly}, and industrial production images~\cite{roth2022towards}. In a recent visual-based blockchain task~\cite{rahouti2022vrchain}, the technology of blockchain was used to provide decentralized communication between robots so they could find their way back home using common visual landmarks. In~\cite{lopes2019controlling}, external parties, Oracle, analyze captured images to determine how many balls need to be picked by UR3 arm. With this information, smart contracts can securely control robots, ensuring that no one can change the logic once it is on the blockchain. In another study~\cite{lopes2019robot}, the authors proposed an approach for securing the robot's workspace and controlling its action using smart contracts and 3D image analysis.

%%%%%%%%%%%%%%%%%%%%%%%%%%%%%%%%%%%%%%%%%%%%%%
%%                                          %%
%%        PROBLEM DEFINITION                %%
%%                                          %%
%%%%%%%%%%%%%%%%%%%%%%%%%%%%%%%%%%%%%%%%%%%%%%

\section{Background}

Through this section, we provide a general overview of our previous works about the proposed visual anomaly detection framework and the distributed ledger technology solution for multi-agent systems.

\subsection{Deep feature extraction and matching}

% In our recent work~\cite{salimpour2022self}, we proposed a visual anomaly detection approach for detecting changes between image pairs as the visual anomalies. Study's main objective was to propose a general method to find anomalies in an unknown environment without prior knowledge. For detecting anomalous regions and objects in the query images, we utilized the deep learning Mask-RCNN instance segmentation model to group unmatched points extracted by the pre-trained SuperPoint~\cite{detone2018superpoint} and SuperGlue~\cite{sarlin2020superglue} deep learning methods into specific segments. In order to find proper unmatched interest points, a single-shot method was used to calibrate the SuperPoint and SuperGlue methods. In the calibration stage, a single-shot method based on matched points in a few pairs of images with linear shifts was used to find proper thresholds for the detection and matching the interest points. There are some new objects that have not been segmented because they do not belong to any class in the segmentation model, the DBSCAN method was applied to cluster their remaining unmatched points. 
Matching and extracting feature points are critical steps when a different environmental condition, such as lighting and viewpoints, affects image comparison and matching. SuperPoint is a fully convolutional neural network self-supervised feature point extraction. It uses a basic detector called MagicPoint that is pre-trained on a synthetically generated dataset consisting of simple shapes, along with homographic adaptation for more training samples from each image. Therefore, it can detect interest points more sensitively than traditional corner detectors~\cite{detone2018superpoint}. 

After the feature extraction process, various methods can be used to take key points and their descriptors in image pairs and match them with corresponding points. SuperGlue is a feature point matching method based on graph neural networks that show better performance for points extracted by the SuperPoint model. The method is based on two layers, a graph neural network, and an optimal matching layer~\cite{sarlin2020superglue}. Using a differentiable Sinkhorn algorithm, matchable points are efficiently paired, and non-matchable points are rejected.

\subsection{IOTA Smart Contracts}

As a subset of the wider DLT domain, blockchain systems have grown in popularity in a variety of use cases. Through smart contracts, specific tasks can be executed when certain conditions are met in distributed applications. The most commonly used blockchain platform for swarm robotics is Ethereum, which uses Turing-complete smart contracts written in Solidity~\cite{strobel2020blockchain}. Ethereum's intrinsic scalability is the main limitation due to its classical single-chain structure. The Tangle~\cite{popov2018tangle}, a directed acyclic graph (DAG)-based DLT, was introduced to solve some of the fundamental weaknesses in classic blockchain systems. IOTA DLT uses the Tangle as its underlying structure, where transactions are the primary data structures. A graph-based ledger rather than a linear chain, which is the concept behind Tangles, would make it more flexible in terms of network partitioning. Multi-robot systems relying on IOTA are therefore unique from a DLT perspective.

As part of ensuring the Tangle's robustness and security, the IOTA foundation has a centralized coordinator that confirms valid transactions. The second version of IOTA, Shimmer, was launched in order to achieve full decentralization. In this work, we use the Go implementation of Shimmer called GoShimmer. As the data structure in the Tangle is graph-based, the implementation of IOTA's smart contract mechanism was challenging. As a solution, the IOTA foundation introduces the IOTA Smart Contract Platform (ISCP) as a second layer on top of the Tangle.In this second layer, Wasp, which is the implementation of ISCP in Go, creates a chain.

%%%%%%%%%%%%%%%%%%%%%%%%%%%%%%%%%%%%%%%%%%%%%%
%%                                          %%
%%              METHODOLOGY                 %%
%%                                          %%
%%%%%%%%%%%%%%%%%%%%%%%%%%%%%%%%%%%%%%%%%%%%%%

\begin{figure}[t]
    \centering
    \includegraphics[width=\textwidth]{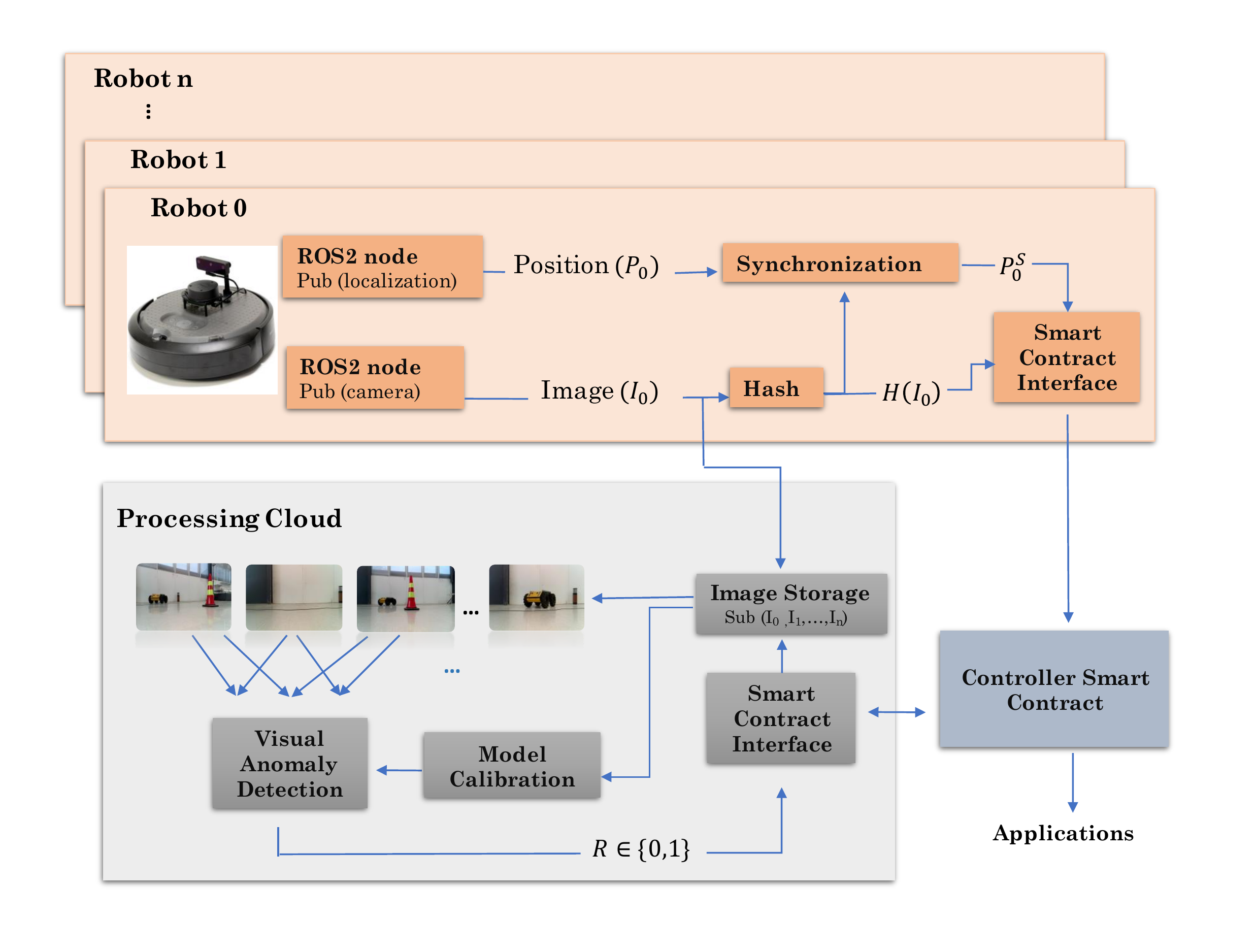}
    \caption{High-level overview of the proposed system architecture. In the current implementation, a trusted processing cloud performs the vision-based anomaly detection. In the future, this workload can be distributed and validated through the smart contracts, but will still run off-chain.}
    \label{fig:diagram}
\end{figure}

% \newpage
\section{Methodology}

Based on the proposed framework for using IOTA smart contracts with ROS\,2 in~\cite{keramat2022partition} for a multi-robot system, we propose the system depicted in~\ref{fig:diagram}. Every robot in this system and the processing cloud runs an instance of GoShimmer and Wasp nodes to form the IOTA network and enable running smart contracts. For simplicity in this proof of concept, the proposed system is not supporting partition tolerance, but according to our earlier framework this can be done easily. All the Wasp nodes create a single chain and deploy the smart contract on it. The smart contract's scheme is shown in~\ref{lst:sc}. The processing cloud is a trusted server to perform the operations which needs high computation power that can not be deployed on the smart contract.

Every robot in the system publishes its position and images captured by the camera. Since the publishing rate of the position and images are different we use a synchronization node in ROS\,2 which checks the timestamp of images and positions and associates a position for each image because the positions are published at a higher rate. Every image is sent to the storage unit of the processing cloud. Then the hash of the image, and the position are submitted to the $submitPair$ function of the smart contract.

The smart contract is mainly responsible to find the pairs of images to be compared and determining which robot is behaving maliciously based on the comparison results. We define intersections for data comparison as the locations where at least $3f+1$ robot have visited  and where images are available with at least a certain overlapping viewpoint (calculated based on position and orientation of the robots). The value $f$ is the number of byzantine robots that our method can tolerate. The schema of the smart contract for IOTA is show in the Listing.~\ref{lst:sc}. The $Pair$ is a struct defined to store the hash of an image, its position, and orientation. Every robot uses the $submitPair$ method to store the images. This function then calls the $findIntersections$ function.

To find the intersections, the $findIntersections$ function divides the map into overlapping $2d\times2d$ cells. The amount of overlap is $d$ among the adjacent cells. In this way every image will be assigned to four adjacent cells based on the location of image. After associating every image to the cells, inside each cell an exhaustive search is performed to find a set of $3f+1$ images each from different robot that has each pair of them has maximum distance of $d$ and maximum orientation difference of $\delta$. By using this method we can make sure that any set of images that have maximum distance of $d$ and maximum orientation difference of $\delta$ will not be missed and it is computationally faster than searching over all the locations since it is related to number of cells instead of number of locations. In each cell, at most one intersection is selected. This also prevents the byzantine robot to compromise from submitting several correct images and one altered image with same position.

The set of images found by $findIntersections$ function is stored in the $sets$ state variable of the smart contract. On the other hand, the processing cloud polls the smart contract by $getIntersection$ functions. The cloud retrieves the set. Based on the hash of images in the set, corresponding images are extracted from the storage unit. Then every two pair of the images are passed to Visual Anomaly Detection module. The result of this module is a binary value indicating if the two images have any difference of not. The cloud submits this results by $submitComparison$ function to the smart contract.

% \begin{minipage}{\linewidth}
\begin{figure}[ht]
\small
\begin{lstlisting}[language=bash,backgroundcolor = \color{white},
caption={Smart Contract Schema.},
frame=single,
label={lst:sc}]
name: VisionByzContract
description: Vision based byzantine detection smart contract
events: {}
structs:
  Pair:
    // Hash of the image and its position
  CompResult:
    // Includes two robot IDs and their score
typedefs: {}
state:
  scores: Int8[]        // Score of each robot
  state: Bool[]         // Indicates if every robot is byzantine or not
  intersections: Pair[] 
  cells: Position[]
funcs:
  init:
    params:
      f: Int32
      n: Int32
  submitPair:
    params:
      pair: Pair
  submitComparison:
    params:
      res: CompResult
  findIntersections:
    access: self
    params: {}
views:
  getIntersection:
    results:
      pairs: Pair[]
  getRobotState:
    results:
      state: Bool
\end{lstlisting}
\end{figure}
% \end{minipage}

The smart contract keeps a score for each robot based on the results of comparisons. A robot's score is incremented by one if the cloud finds a difference in the comparison. If we suppose that robots pass by sufficient amount of intersections, the smart contract can detect the byzantine robot based on the scores. If the robot's score is bigger than the average of scores by a certain threshold, the robot can be marked as byzantine. This decision also stored in the smart contract and can be used for further applications.

\subsection{Visual anomaly detection module}

In our last study~\cite{salimpour2022self}, we proposed a general framework to detect regions that have changed in pair images as pixel-based visual anomalies. To detect anomalies in an unknown environment without training, we applied pre-trained deep learning models for extraction, matching, and segmentation. The performance and final results of SuperPoint and SuperGlue are influenced by confidence parameters such as keypoint detection and matching confidence thresholds. We applied a few-shot calibration procedure based on the coefficient of variation of matched keypoints to find the optimum matching and extracting thresholds. 

In order to find overlap areas, we apply masks to both images based on matched interest points. Then, we segment not matched points using the Mask-RCNN instance segmentation method. As the segmentation model is not trained for all objects, we use DBSCAN clustering algorithm to group the remaining not-matched points belonging to new foreign objects. The proposed system architecture for visual anomaly processing is illustrated in Fig~\ref{fig:diagram} in the processing clould unit. When a change is detected in either image of an image pair, the processing cloud returns True for both suspicious images, otherwise it returns False. With this, we measure compliance or disparity between pairs of images, and allows us to build a measure of trust within the system.

%%%%%%%%%%%%%%%%%%%%%%%%%%%%%%%%%%%%%%%%%%%%%%
%%                                          %%
%%              EXPERIMENTS                 %%
%%                                          %%
%%%%%%%%%%%%%%%%%%%%%%%%%%%%%%%%%%%%%%%%%%%%%%

\section{Experimental Results}

This section discusses the results of the byzantine robot detection experiment with ground robots that was conducted to evaluate the functionality and effectiveness of the proposed framework.

\begin{figure}[t]
    \centering
    \setlength{\figurewidth}{0.88\textwidth}
    \setlength{\figureheight}{0.66\textwidth}
    \input{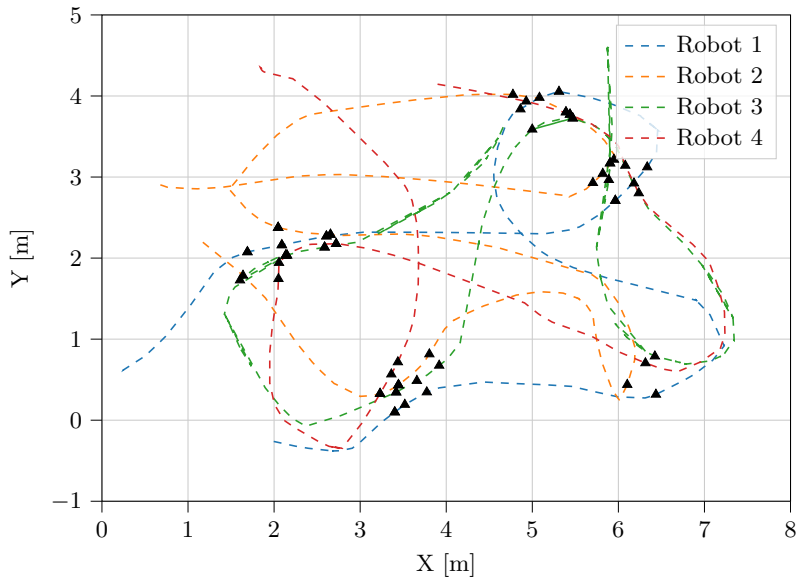}
    \caption{Trajectories of the different robots during the experiments. The yellow circles show the locations that the smart contract computes for data comparisons.}
    \label{fig:trajectory}
\end{figure}

\begin{figure}[t]
    \centering
    \setlength{\figureheight}{0.55\textwidth}
    \setlength{\figurewidth}{0.88\textwidth} 
    \scriptsize{\input{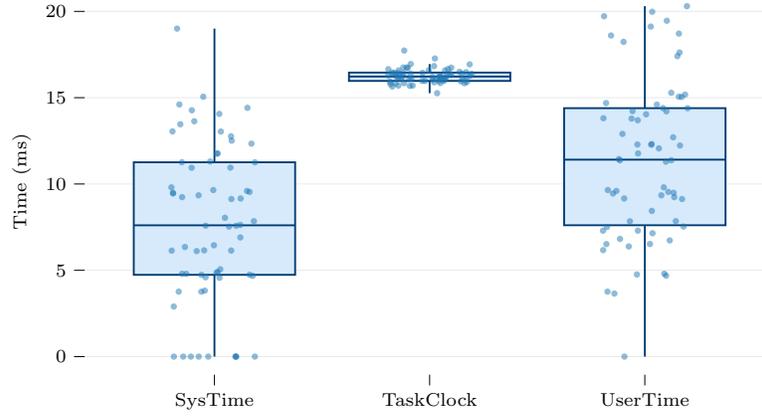}}
    \caption{Analysis of the CPU utilization time of the $submitPair$ method measured with the Linux perf tool. The graph demonstrates a low performance impact induced by the blockchain integration.}
    \label{fig:sc_exec_time}
\end{figure}

\subsection{Experiment setup}

\textit{Hardware.} The employed multi-robot system in this paper consists of four ground TurtleBot4 Lite robots built on top of the iRobot%\textcircled{R} 
Create 3 mobile base with a Luxonis OAK-D-Lite stereo camera. For localization, we utilize an external motion capture system with four Optitrack PrimeX 22 cameras and robots move in the area defined by the motion capture system forming a square of approximately 48\,$m^2$. In this study, we placed several objects from a variety of classes of the COCO dataset in the environment, along with some unknown objects and some with unclear textures.

\textit{Software.} The Turtlebot4 robots run ROS\,2 Galactic under Ubuntu 20.04 and publish camera images at 1\,Hz. Localization is running in ROS\,1 Noetic with a 120\,Hz publishing rate. Data is forwarded from ROS\,1 to ROS\,2 with a the ros1\_bridge package and the data from the two topics synchronized for saving image locations. Each robot explores the environment following a predefined trajectory and publishes its topics to the trusted storage unit and through the smart contract interface. Figure~\ref{fig:trajectory} illustrates the trajectory of the robots during the experiment.

The trusted processing cloud is represented in the experiments by a computer with an Intel Core i7-11800H processor and 64\,GB of memory. Robots and the processing unit are running Wasp v0.2.5 and GoShimmer v0.7.4 nodes and they are all connected to the same wireless network. The smart contract schema illustrated in Listing.~\ref{lst:sc} is implemented in the Go programming language.

\begin{figure}[t]
    \centering
    \includegraphics[width=\textwidth]{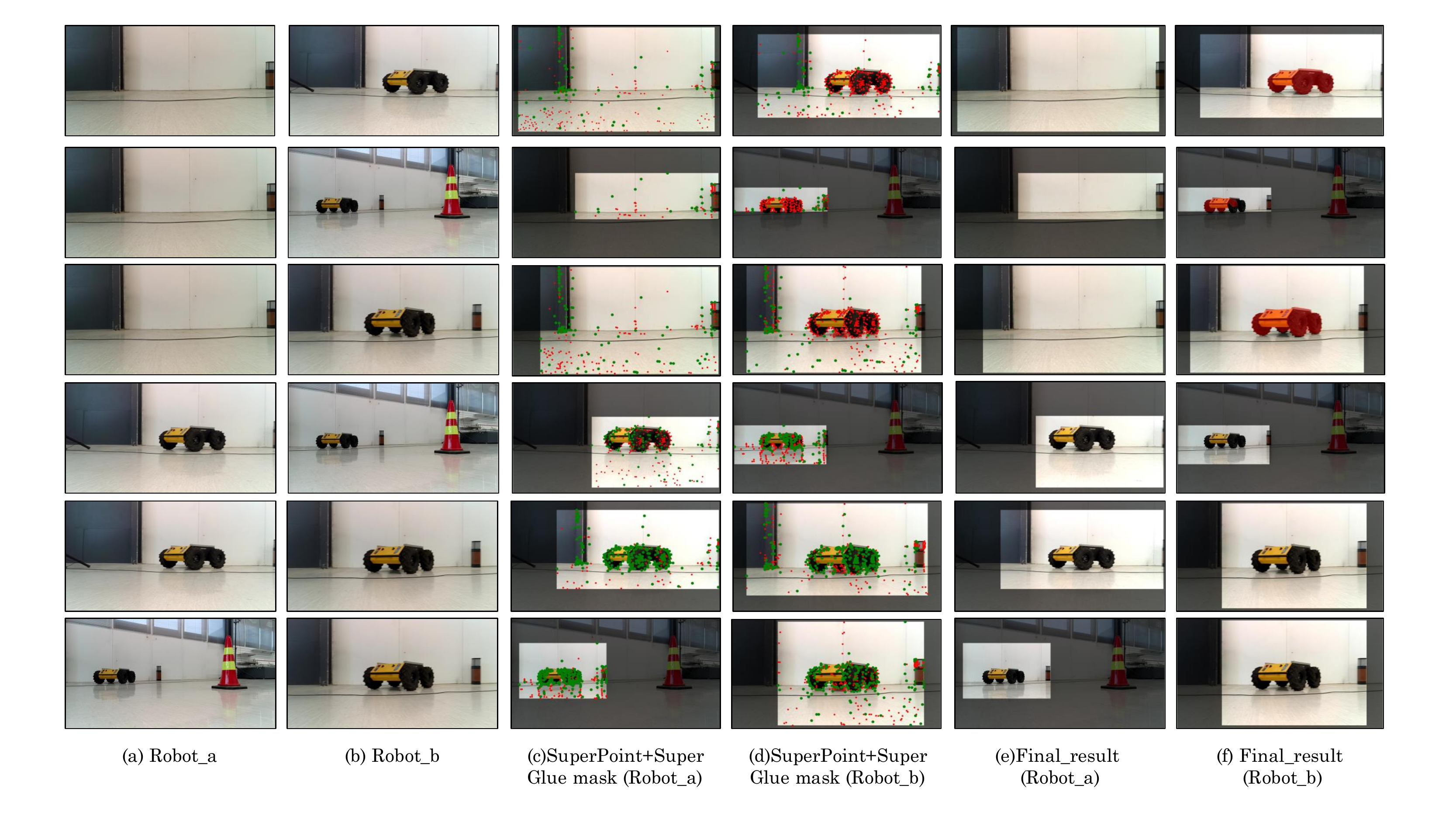}
    \caption{Comparison of pair images related to the required locations by IOTA smart contract. The first two columns represent images submitted by robots. After feature extraction and clustering, the final results represent different robot pairs.}
    \label{fig:iota_6pairs}
\end{figure}

\subsection{Smart Contract}

In this experiment, we choose $f=1$ to tolerate at most one byzantine robot. To find an intersection $3f+1=4$ robot should therefore visit the same location with similar orientation to allow for enough overlapping pixel area in the images. In the smart contract we set $d=0.5m$ and $\delta=0.4 \text{rad}$ to define an intersection. In Fig.~\ref{fig:trajectory}, we illustrate the path traversed by each robot. The smart contract outputs 17 intersection sets. All points in these sets are marked by black triangles in the figure. The CPU utilization time is illustrated in Fig.~\ref{fig:sc_exec_time}, measured by the Linux perf tool.

\begin{figure}[t]
    \centering
    \includegraphics[width=\textwidth]{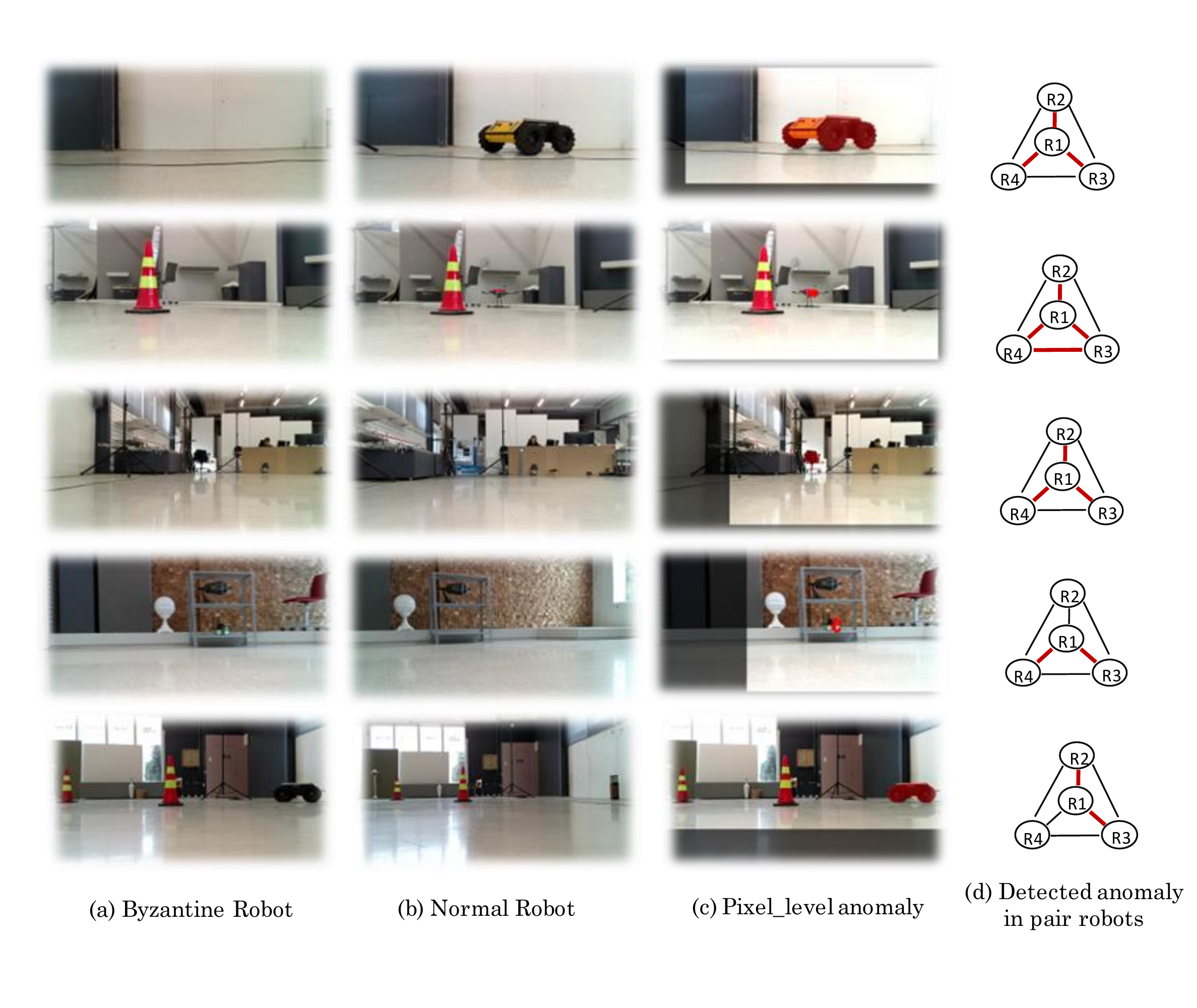} \\
    \vspace{-2.5em}
    \caption{The image submitted by the byzantine robot (a) and the image submitted by a honest robot (b). In column (c) the anomalies detected by the visual anomaly detection module are marked in red. And column (d) illustrates how this comparisons effect the score of each robot.}
    \label{fig:2pairs_iota}
\end{figure}

\begin{figure}[t]
    \centering
    \setlength{\figurewidth}{.88\textwidth}
    \setlength{\figureheight}{.55\textwidth}
    % This file was created with tikzplotlib v0.10.1.
\begin{tikzpicture}

\definecolor{crimson2143940}{RGB}{214,39,40}
\definecolor{darkgray176}{RGB}{176,176,176}
\definecolor{darkorange25512714}{RGB}{255,127,14}
\definecolor{forestgreen4416044}{RGB}{44,160,44}
\definecolor{lightgray204}{RGB}{204,204,204}
\definecolor{mediumpurple148103189}{RGB}{148,103,189}
\definecolor{steelblue31119180}{RGB}{31,119,180}

\begin{axis}[
    width=\figurewidth,
    height=\figureheight,
    legend cell align={left},
    legend style={
      fill opacity=0.8,
      draw opacity=1,
      text opacity=1,
      at={(0.03,0.97)},
      anchor=north west,
      draw=lightgray204
    },
    tick align=outside,
    tick pos=left,
    x grid style={darkgray176},
    xlabel={t [s]},
    xmin=55.68, xmax=150.72,
    xtick style={color=black},
    y grid style={darkgray176},
    ylabel={Score},
    ymin=-1.1, ymax=45.1,
    ytick style={color=black}
]
\addplot [semithick, steelblue31119180, const plot mark right]
table {%
60 3
61.2 5
63.6 8
65.4 10
90 13
91.8 16
92.4 18
94.2 21
102 23
103.2 25
105 27
105.6 29
120.6 32
144 35
145.2 38
145.8 40
146.4 43
};
\addlegendentry{Robot1}
\addplot [semithick, darkorange25512714, const plot mark right]
table {%
60 1
61.2 2
63.6 3
65.4 4
90 5
91.8 7
92.4 8
94.2 9
102 10
103.2 11
105 13
105.6 14
120.6 15
144 17
145.2 18
145.8 18
146.4 19
};
\addlegendentry{Robot2}
\addplot [semithick, forestgreen4416044, const plot mark right]
table {%
60 1
61.2 2
63.6 3
65.4 5
90 6
91.8 7
92.4 9
94.2 10
102 10
103.2 12
105 13
105.6 14
120.6 15
144 17
145.2 18
145.8 19
146.4 20
};
\addlegendentry{Robot3}
\addplot [semithick, crimson2143940, const plot mark right]
table {%
60 1
61.2 1
63.6 2
65.4 3
90 4
91.8 6
92.4 7
94.2 8
102 9
103.2 10
105 11
105.6 11
120.6 12
144 13
145.2 14
145.8 15
146.4 16
};
\addlegendentry{Robot4}
\addplot [semithick, mediumpurple148103189, const plot mark right, dashed]
table {%
60 1.995
61.2 3.325
63.6 5.32
65.4 7.315
90 9.31
91.8 11.97
92.4 13.965
94.2 15.96
102 17.29
103.2 19.285
105 21.28
105.6 22.61
120.6 24.605
144 27.265
145.2 29.26
145.8 30.59
146.4 32.585
};
\addlegendentry{Threshold}
\end{axis}

\end{tikzpicture}\\
    \vspace{-1em}
    \caption{Accumulated score of each robot over time and the threshold determining the byzantine robot. Through the mission, Robot\,1 is deemed byzantine having too high disparity score.}
    \label{fig:score}
\end{figure}
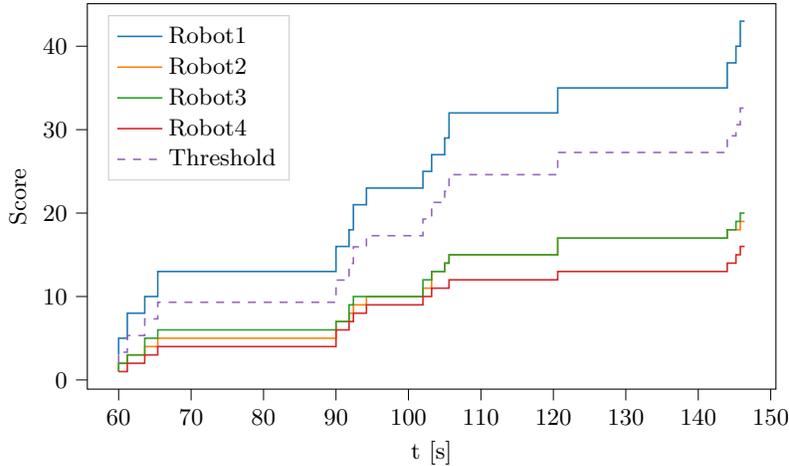

\vspace{1em}
\textit{Calibration.} 

The image storage section chooses a few image pairs of the operational environment with the linear shift to calculate matching and extracting thresholds in order to calibrate the SuperPoint + SuperGlue model. The optimum matching threshold is $\Delta=0.35$, and the best keypoint extracting threshold is $\lambda = 0.001$. In both cases the selected value differs from the default value of $\Delta=0.2$ and $\lambda = 0.005$.

\vspace{1em}
\textit{Anomalous object identification.} 

We now describe experiments we conducted with the ground robots to demonstrate the effectiveness of the proposed byzantine agent detection process based on the results of visual anomaly detection.

Figure~\ref{fig:iota_6pairs} illustrates the result of processing a set of images meeting the requirements in terms of relative position and orientation. These comparisons are requested by the smart contract. A two-by-two analysis is conducted for images from the four robots. As can be seen in the figure, the proposed model identifies three anomalies related to robot number\,1, and one related to robot number\,4, over a total of six comparisons.

Figure~\ref{fig:2pairs_iota} shows a series of images received from a byzantine robot and a \textit{normal} robot in different places. In inspection or monitoring applications, the anomaly detected during the visual processing stage could be an object which is removed or added in the time elapsing between the visits of the different robots to that location. Alternatively, the same approach serves to identify altered or manipulated data from potentially malicious agents. Our method is robust to different classes, so it is capable of clustering non-matched points when they cannot be segmented by the Mask-RCNN model based on the predefined set of known objects. There main limitations are, however, with some new texture-less objects that cannot be detected because of the lack of enough interest points to match.

In Figure.~\ref{fig:2pairs_iota} Robot\,1 is the byzantine robot. The graph $G=(V, E)$ in each row illustrates the comparisons between each pair of images. Vertices of the graph represents each robot, and edges represents the result of the comparison between the images from the corresponding vertices. Red edges indicate that there is an anomaly detected in the comparisons. For every set score of a robot is calculated by summing the number of red edges connected to its vertex. For example in the first row, the score of Robot\,1 is 3 and the score of Robot\,3 is 1. Figure.~\ref{fig:score} illustrates the accumulated score of each robot over the time. The threshold is defined 30 percent bigger than the average of all scores.

In summary, our method is able to effectively identify anomalies or changes and build a measure of trust (through measuring similarities and disparities between images submitted by different robots) within the system. Even though this proof of concept relies on a central trusted server for the actual data comparisons, the smart contracts deciding what data is to be compared run in a decentralized manner, and the results are available through the system with the smart contract interface. The types of anomalies or changes identified do not necessarily imply malicious or byzantine behaviour, but rather flag robots that gather data that differs more from the rest of the fleet. In a practical scenario, a human operator would analyze the data whenever a robot is flagged as potentially byzantine.

%%%%%%%%%%%%%%%%%%%%%%%%%%%%%%%%%%%%%%%%%%%%%%
%%                                          %%
%%              CONCLUSION                  %%
%%                                          %%
%%%%%%%%%%%%%%%%%%%%%%%%%%%%%%%%%%%%%%%%%%%%%%

% \newpage
\section{Conclusion}
\label{sec:conclusion}

This paper presents a solution to byzantine agent detection using IOTA smart contracts. In comparison with the state-of-the-art, with solutions that are mostly based on the Ethereum blockchain, the IOTA-based approach presented is theoretically more scalable, has lower computational impact and can be implemented to be partition tolerant. Additionally, we show the integration of more complex data comparison using deep learning and a vision-based approach to byzantine agent detection. Our results show that the blockchain layer adds negligible computational overhead, while the anomaly detection algorithm allows for building a measure of trust among the robots in the system. Our system effectively detects a potentially byzantine robot based on the disparity of its data when compared to the data gathered by other robots. In future works, we aim at fully decentralizing the proposed system by adding incentives for robots or other nodes in the blockchain to run the deep learning inference models, so that their result can also be validated even when part of the data processing occurs off-chain.

%%%%%%%%%%%%%%%%%%%%%%%%%%%%%%%%%%%%%%%%%%%%%%
%%                                          %%
%%            ACKNOWLEDGMENT                %%
%%                                          %%
%%%%%%%%%%%%%%%%%%%%%%%%%%%%%%%%%%%%%%%%%%%%%%

\section*{Acknowledgment}

This research work is supported by the R3Swarms project funded by the Secure Systems Research Center (SSRC), Technology Innovation Institute (TII).

%%%%%%%%%%%%%%%%%%%%%%%%%%%%%%%%%%%%%%%%%%%%%%
%%                                          %%
%%              BIBLIOGRAPHY                %%
%%                                          %%
%%%%%%%%%%%%%%%%%%%%%%%%%%%%%%%%%%%%%%%%%%%%%%
% \newpage
% \nocite{*}
\bibliographystyle{splncs04}
\bibliography{bibliography}

\end{document}